\documentclass[10pt,twocolumn,letterpaper]{article}

\usepackage{iccv}
\usepackage{times,booktabs,multirow}
\usepackage{epsfig}
\usepackage{graphicx}
\usepackage{amsmath}
\usepackage{amssymb}

\usepackage{gensymb}

\usepackage[pagebackref=true,breaklinks=true,letterpaper=true,colorlinks,bookmarks=false]{hyperref}
\usepackage{authblk}
 \iccvfinalcopy 

\def\iccvPaperID{2637} 

\ificcvfinal\fi
\begin{document}

\title{Deep Meta Functionals for Shape Representation\vspace{-.75cm}}

\author[1]{Gidi Littwin}
\author[1,2]{Lior Wolf}
\affil[1]{Tel Aviv University} 
\affil[2]{Facebook AI Research}

\maketitle

\begin{abstract}

We present a new method for 3D shape reconstruction from a single image, in which a deep neural network directly maps an image to a vector of network weights.  The network \textcolor{black}{parametrized by} these weights represents a 3D shape by classifying every point in the volume as either within or outside the shape. The new representation has virtually unlimited capacity and resolution, and can have an arbitrary topology. Our experiments show that it leads to more accurate shape inference from a 2D projection than the existing methods, including voxel-, silhouette-, and mesh-based methods. The code is available at: \url{https://github.com/gidilittwin/Deep-Meta}.

\end{abstract}

\section{Introduction}

We propose a novel deep learning method for representing shape and for recovering that representation from a single input image. Every shape is represented as a deep neural network classifier $g$, which takes as input points in 3D space. In addition, the parameters (weights) of the network $g$ are inferred from the input image, by another network $f$.

The method is elegant and enables an end-to-end training with a single and straightforward loss. As a level set surface representation, one is guaranteed to obtain a continuous manifold. Since every point in 3D is assigned a value by $g$, efficient (and even differentiable) rendering is obtained. For the same reason, unlike voxel or point-cloud based methods, the gradient information is given in every point in 3D, making training efficient. This gradient information, however, is more informative near the shape's boundary. Therefore, we propose a simple scheme of selectively sampling 3D points during training, such that points near the objects boundary are over-represented.

In contrast to most other methods, which suffer from a capacity limitation, the capacity of the 3D surface is exponential in the number of parameters of network $g$. Even for relatively small networks, it exceeds what is required by all graphical applications.

In contrast to mesh based methods, the topology of the resulting shape is not limited to a template shape and it can have an arbitrary topological complexity.

Our experiments show that in addition to these modeling and structural advantages, the method also results in better benchmark performance than the existing ones.

\section{Previous work}

\begin{table*}[t]
\begin{center}
\begin{small}
\begin{tabular}{@{}llllll@{}}
\toprule
 & \textbf{Voxels} & \textbf{Point Clouds} & \textbf{Polygon Mesh} & \textbf{Implicit functions} & \textbf{Meta Functionals} \\
\midrule
Memory Footprint & High* & Low & Low & High & Low \\
Reconstruction Resolution & Limited by memory & High & Limited by template mesh & Unlimited & Unlimited\\
Topology & Limited by resolution & No topology & Limited by template mesh & Unlimited & Unlimited \\
Train Time & Long & Short & Short & Long & Short \\
Rendering & Suited & Suited & Very suited & Suited & Suited\\
\bottomrule
\end{tabular}
\end{small}
\end{center}
\caption{Comparison of the major traits between prominent 3D representation approaches. *The memory footprint of voxel representation has been somewhat alleviated by more elaborate hierarchical data structures.  }
\label{tab:compare}
\end{table*}

Propelled by the availability of large scale CAD collections such as ShapeNet~\cite{chang2015shapenet} and the increase in GPU parallel computing capabilities, learning based solutions have become the method of choice for reconstructing 3D shapes from single images. 
Generally speaking, the 3D representations currently in use fall into three main categories: (i) grid based methods, such as voxel, which are 3D extensions of Pixels, (ii) topology preserving geometric methods, such as polygon meshes, and (iii)  un-ordered geometric structures such as point clouds. 

Grid based methods form the largest body of work in the current literature. Voxels, however, do not scale well, due to their cubic memory to resolution ratio. To address this issue, researchers have come up with more efficient memory structures. Riegler et al.~\cite{Riegler_2017} , Tatarchenko et al.~\cite{tatarchenko2017octree} and H{\"a}ne et al.~\cite{hane2017hierarchical} use nested tree structures (Octrees) to leverage the inherent sparsity of the voxel representation. Richter et al.~\cite{richter2018matryoshka} introduce an encoder decoder architecture, which decodes into 2D nested shape layers that enable reconstruction of the 3D shape.  

A different approach for handling the inherent sparsity of the data is using a point cloud representation. Point clouds form an efficient and scaleable representation. Fan et al.~\cite{Fan_2017_CVPR} designed a point set generation network, which Jiang et al.~\cite{Jiang_2018_ECCV} improved, by adding a geometric consistency loss via re-projected silhouettes and a point-based adversarial loss. The clear disadvantage of this approach is the ambiguous topology, which needs to be recovered in post-processing, in order for the object to be properly lit and textured.

Another form of 3D representation that is especially suited for 2D projections is the polygon mesh. Kato et al.~\cite{kato2018neural} introduced a render-and-compare based architecture that enables back-propagation of gradients, through a 2D projection of a template mesh. In order to facilitate meaningful training, they designed a differentiable mesh rendering pipeline that approximates the gradients of a silhouette-comparing cost function. Liu1 et al.~\cite{liu2019soft} extended their work, by designing a more efficient differentiable rendering engine to produce very compelling results. Wang et al.~\cite{wang2018pixel2mesh} employed an innovative graph based CNN to extract perceptual features from an image, utilizing a pre-trained VGG network in a fully supervised scenario.

There are some works that break from these categories. Groueix et al.~\cite{Groueix_2018} learn to generate a surface of a 3D shape by predicting a collection of local 2-manifolds and obtaining the global surface by applying a union operation. 

\textcolor{black}{Recently and concurrently with our work, several publications demonstrated the usage of continues implicit fields for shape representation. Chen et al.~\cite{chen2018learning}, Park et al.~\cite{park2019deepsdf} and Mescheder et al.~\cite{mescheder2018occupancy} used an MLP conditioned on a shape embedding to represent shapes. While the authors used slightly different formulations and conditioning techniques to achieve the goal of shape representation, the common attribute to all three methods is a large MLP that acts as a decoder. Contrary to these methods, our decoder decodes the embedding vector into a set of weights which parameterize a function space that in turn, forms a mapping between samples in space and shape occupancy. At train and inference time, the model generates decoders that are uniquely defined for each shape and so are very parameter efficient.}

These outlined categories for 3D representations all suffer from different drawbacks and present different advantages, see Tab.~\ref{tab:compare}. Grid based approaches draw from a large body of work conducted in parallel topics of research but do not scale well or require elaborate custom layers to handle these restrictions. Point cloud based methods overcome this limitation but do not reconstruct topologically coherent shapes or require post-processing to do so. Polygon mesh based methods are more suited in nature for 2D supervision but enforce a very restrictive representation, which prevents reconstruction of even very simple shapes that exhibit different topology than the chosen template. \textcolor{black}{The recently introduced implicit shape based methods~\cite{chen2018learning,park2019deepsdf,mescheder2018occupancy} overcome most of these issues but pay a price in the form of very long train times (as reported by the authors) and a very large decoder which is problematic when evaluating in high resolution. It is also not clear how these methods generalize to very large training sets which include multiple shape classes since none of these publications have reported results on the commonly used ShapNet ground-truth annotations and instead opted with retraining the baseline methods on subsets of the data.  Mescheder et al.~\cite{mescheder2018occupancy} is the only implicit-shape method to report multi-class results, but introduced additional supervision in the form of a pre-trained on imagenet.} 

\smallskip
\noindent{\bf Implicit Surfaces} The classical active contour methods, first introduced by Kass \etal~\cite{kass1988snakes}, have employed  energy-minimizing iterations to guide an image curve (also known as a snake) towards image features, such as image edges. Limited in topology and suffering from an ineffective evolution procedure, the method was reformulated as a level set method~\cite{caselles1993geometric, caselles1997geodesic, malladi1994evolutionary,kichenassamy1995gradient}. The level set method was generalized to volumetric 3D data~\cite{krueger2008active}. The literature level set methods are mostly used for evolving a curve. This scenario is vastly different than our method, which uses the level set of a classifier at the natural threshold of 0.5, and employs a direct regression for obtaining the parameters of that classifier. The properties of the level set representation still carry over to our case.

\smallskip
\noindent{\bf Hypernetworks or dynamic networks} refer to a technique in which one network $f$ is trained to predict the weights of another network $g$. The first contributions learned specific layers for tasks that require an adaptive behavior~\cite{klein2015dynamic,7410424}. Fuller dynamic networks were subsequent used for video frame prediction~\cite{jia2016dynamic}. The term hypernetwork is due to~\cite{ha2016hypernetworks}, and the application to few-shot learning was introduced in~\cite{bertinetto2016learning}.

\section{Method}

The method employs two networks $f,g$ with parameter values $\theta_f,\theta_I$ respectively. The network weights $\theta_f$ are fixed in the model and are learned during the training phase. The weights of network $g$ are a function of input image $I$, given as the output of the network $f$.  

The two networks represent different levels of the shape abstraction. $f$ is a mapping from the input image $I$ to the parameters $\theta_I$ of network $g$, and $g$ is a classification function that maps a point $p$ with coordinates $(x,y,z)$ in 3D into a score $s_I^p\in [0,1]$, such that the shape is defined by the classifier's decision boundary.

The model is formally given by the following equations:
\begin{align}
    \theta_I = f(I,\theta_f)\\
    s_I^p = g(p,\theta_I) \label{eq:g}
\end{align}
We parameterize $f(I,\theta_f)$ as a CNN and $g(p,\theta_I)$ as a Multi-Layered Perceptron (MLP). 
A-priori, it is not clear that a generic architecture for $g$ can perform the modeling task. The normalized shapes in the ShapeNet dataset represent closed 2D manifolds restricted to the 3D cube $x,y,z\in\{-1,1\}$. $g(p,\theta_I)$ should be able to accurately capture both inter and intra shape variations. As we show in our experiments, a fully connected neural network with as few as four hidden layers and less than 5000 trainable parameters is indeed an adequate choice.

Training is done with a single loss, which is the cross-entropy classification loss. Let the score $s_I^p\in \mathbb{R}$ represent a Bernoulli distributions $[1-g(p,\theta_I),g(p,\theta_I)]$ and let $y(p)\in\{0,1\}$ be the ground truth target representing whether the point $p$ is inside ($y(p)=1$) or outside ($y(p)=0)$ the shape. 

The unweighted loss of the learned parameters $\theta_f$, for image $I$ with ground truth shape $y$ is given by
\begin{multline}
H(\theta_f,I) = -\int_V y(p)log(g(p,f(I,\theta_f))) +\\ (1-y(p))log(1-g(p,f(I,\theta_f)))dp 
\end{multline}
where $V$ is the 3D volume in which the shapes reside. During training, the integral is estimated by sampling points in the volume $V$.

\paragraph{Point sampling during training}

Similar to the training of other classifiers, the points near the decision boundary are more informative. Therefore, in order to make the training more efficient, we sample more points in the vicinity of the shape's boundary. 

This sampling takes place in the vicinity of every vertex of the ground truth mesh. A uniform Gaussian with a variance of $0.1$ is used. The label is computed efficiently, by using a voxel occupancy grid for each shape.  

\begin{figure}
\centering
\includegraphics[width=0.9291\linewidth]{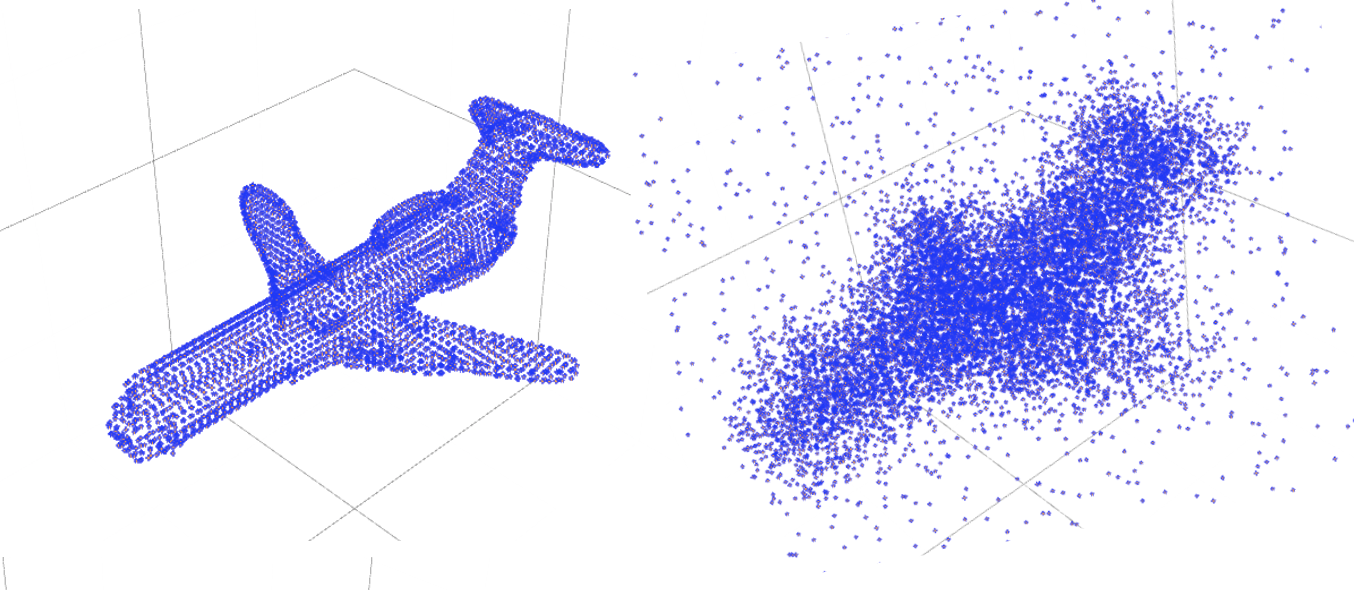}
\caption{Point sampling. On the left, the mesh vertices and on the right points sampled during training.  }
\label{fig:sample}
\end{figure}

At every training batch, we sample a fixed number of points from every shape sample in the batch. In order to cover regions of space that are scarcely sampled due to the shape distribution, we add 10\% of uniformly distributed points to each sample. See Fig.~\ref{fig:sample} for an illustration.

\paragraph{Architecture}
\label{sec:arch}

\begin{figure}
\centering
\includegraphics[width=.910\linewidth]{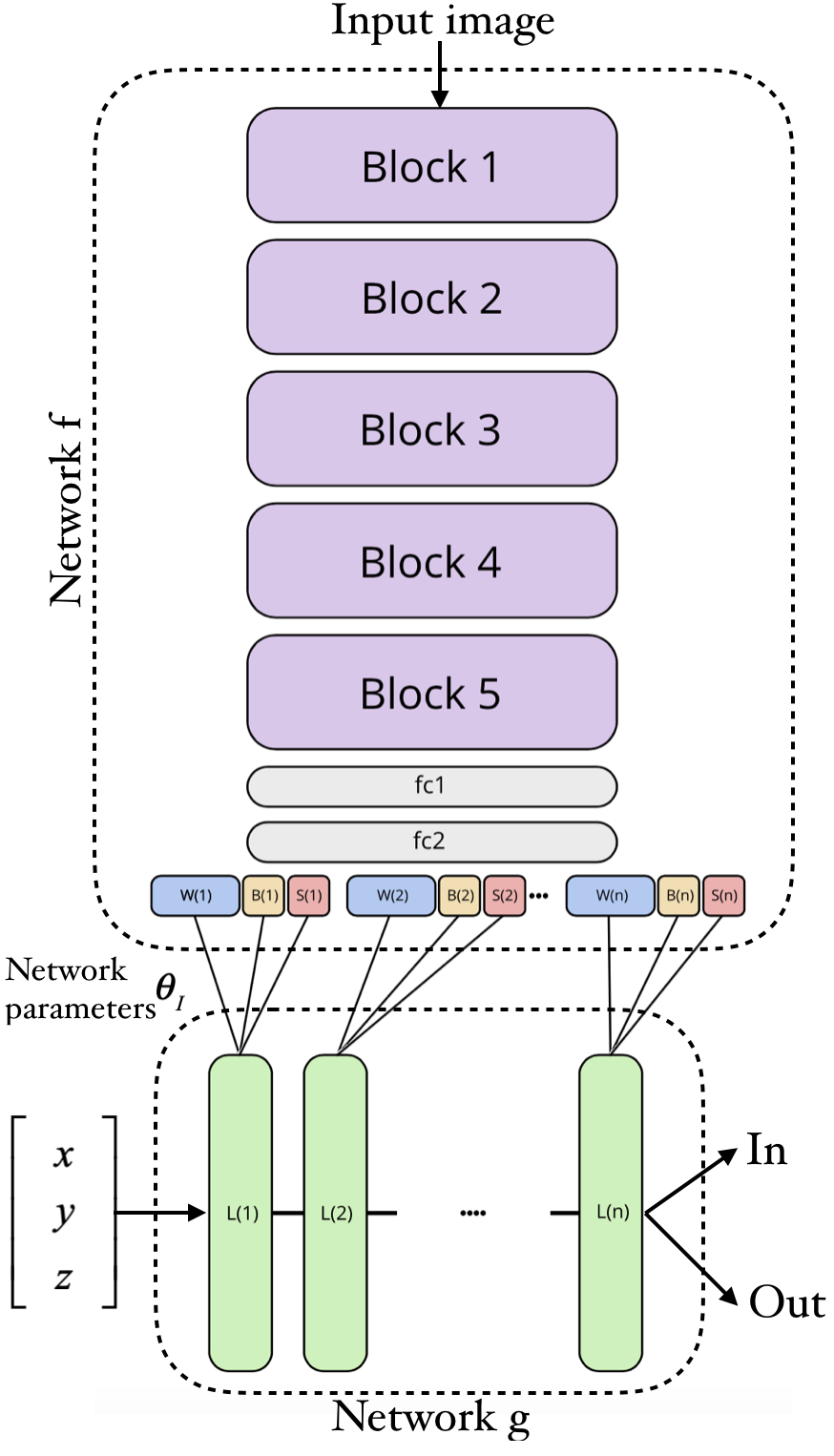}
\caption{The architecture of our neural networks. $f$'s output given an input image $I$ is the set of parameters $\theta_I$ of the network $g$. These include weights, bias, and scale parameters. The network $g$ classifies each input point as either inside or outside the object.}
\label{fig:arch}
\end{figure}

The architecture of the networks is depicted in Fig.~\ref{fig:arch}. Network $f$ is a ResNet with five blocks; $g$ is a fully connected.

The network $g(p,\theta_I)$ is an MLP which maps points $p\in \mathbb{R}^3$ to a scalar field. Our default architecture includes four hidden layers with 32 neurons per hidden layer. In order to make this architecture more suitable for regression, we add a scaling factor that is separate from the weight matrix. Each layer $n$ performs the following computation:
\begin{equation}
y = ((\theta_I^{W(n)} x)\cdot \theta_I^{s(n)} ) + \theta_I^{b(n)}     
\end{equation}
where $x$ is the layer's input, $y$ is its computation result, $\theta_I^{W(n)}$ is the weight matrix of layer $n$, $\theta_I^{b(n)}$ is the bias vector of that layer, and $\theta_I^{s(n)}$ is the learned scale vector. The multiplication between the weighted input and the scale vector is done per coordinate. 

For the network $g$, the ELU activation function~\cite{clevert2015fast} is used. However, the experiments reveal that ReLU or $tanh$ are almost as effective.

Note that the weights of network $g$ are, in fact, feature maps produced by network $f$ and, therefore, represent a space of functions constrained by the architecture of $g$. \textcolor{black}{The architecture presented includes $3394$ parameters and so is very efficient for both training and inference.}

 $f(I,\theta_f)$ is a ResNet very similar in structure to the ResNet-34 model introduced by He et al.~\cite{He:2015:DDR:2919332.2919814}. It starts with a convolutional layer that operates on $I$ with $N$ $(5\times5)$ kernels and then goes through $B$ consecutive blocks, which share the same structure. 
 
 Each one of the blocks is comprised of 3 residual modules, all utilizing $(3\times3)$ kernels. The first residual module in each block reduces the spatial resolution by 2 via strided convolutions and increases the number of feature maps by 2. The succeeding modules keep both spatial and feature dimensionalities. The modules use the pre-activation scheme (BN-ReLU-Conv). The network then employs an average pooling layer, which yields a feature vector of size $(16\times N)$. $K$ fully connected layers with $(16\times N)$ neurons each are applied to this feature vector (ReLU-Conv-Relu-Conv for $K=2$). This results in a feature vector of size $(16\times N)$, which we view as the shape embedding $e(I,\theta_f)$.
 
The $f$ network then splits into multiple heads. There is one group of heads per each layer of $g$, indexed by $n=1,2,...L$,  and each group contains a set of linear regressors that provide the weights for this layer (a matrix $\theta_I^{W(n)}$), the bias term (a vector  $\theta_I^{b(n)}$), and the scale vector ($\theta_I^s(n)$).

Unless otherwise specified, we use $N=64$, $B=5$, $K=2$, and $L=4$. However, as our experiments show, the performance is stable with regards to these parameters.

\paragraph{Rendering}

Since we wish to use off the shelf renderers, rendering is done via the following procedure; see Sec.±\ref{sec:discussion} for a discussion of future renderers. First, we evaluate the field $s_I^p=$ (Eq.~\ref{eq:g}) using a grid of points $p\in [-1,1]^3$ with a spatial resolution of $128$ in each axis. The marching cube algorithm~\cite{Lorensen:1987:MCH:37402.37422} is then applied to obtain a polygon mesh. 

Note that the rendering resolution is not limited to the resolution used in training and in-fact, is only limited by computing resources.

\section{Properties of the representation}

The shape is defined by the isosurface of $g$ at the level of 0.5. Since $g$ employs ELU activation units, it is differentiable. Therefore, by using known results for level sets, from the implicit function theorem, the obtained surface is a smooth manifold~\cite{kosinski2007differential}. This property is obtained, without restricting to a certain mesh topology, unlike other methods.

In order to understand the capacity of the shape defined by $g$, we consider the equivalent network, where the ELU activations are replaced by ReLU ones. For such a network, the number of linear regions is upper bounded by $O((\frac{n}{n_0})^{(L-1)n_0}n^{n_0})$ for a network with $n_0$ inputs, $L$ hidden layers and $n>n_0$ neurons per hidden layer~\cite{montufar2014number}. For the architecture of network $g$, this amounts to between $1\mathrm{e}{+4}$ to $8.6\mathrm{e}{+19}$ linear regions for our smallest MLP (three layers with 16 hidden units each) and our largest tested MLP (six layers with 64 hidden units) respectively. While only a subset of these regions are included in the decision boundary itself, it demonstrates that a network-based representation can present a very high shape representation capacity, even for relatively shallow and narrow networks. This capacity increases exponentially in $L$ and polynomially in $n$.



\section{Experiments}

\begin{figure}
\centering
\hspace{-1mm}\includegraphics[width=.95810\linewidth]{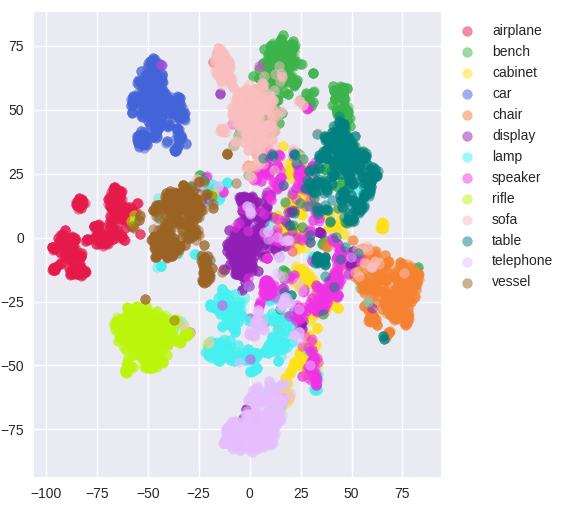}
\caption{A t-SNE visualizations of object embedding from the 13 main categories of the ShapeNet-Core V1 test set }
\label{fig:tsne}
\end{figure}

We demonstrate the effectiveness of our method by comparing it to other state-of-the-art-methods. \textcolor{black}{Experiments are conducted on 2 base resolutions of $32^3$ and $256^3$. For the low resolution experiments} we use the dataset provided by Choy et al.~\cite{choy20163d}, which includes more than 40k objects spanning 13 categories. Each object is rendered from 24 different views sampled uniformly but with a fixed elevation axis viewpoint of {$30\degree$}. The image resolution is set to $(137\times137)$ and the voxel grid resolution is set to 32 on each axis. This resolution limits the resolution of the network's output. However, it allows a direct comparison with previous work. For a fair comparison, we also use the same train/test split used by the authors. \textcolor{black}{For the high resolution experiments, we used the data provided by H{\"a}nee et al~\cite{hane2017hierarchical}, which introduced higher quality rendered images at the resolution of $(224\times224)$ that were sampled at a wider elevation angle distribution of {$-20\degree:30\degree$}. The dataset, as provided by the authors, is generated in two grid resolutions of $32^3$ and $256^3$ and split into train/validation/test sets.}

\subsection{Training and qualitative results}

The network, with a shape parameters of $N=64, B=5, K=2$ and $L=4$ (Sec.~\ref{sec:arch}) was trained for 20 epochs (around 4 days), starting with a learning rate of $5\mathrm{e}{-5}$, and reducing by a factor of 10 after 10 epochs and by a factor of two after 5 additional epochs. One network was trained for all classes, without enjoying the class information. 

\begin{figure}[t]
\centering
\includegraphics[width=.95\linewidth]{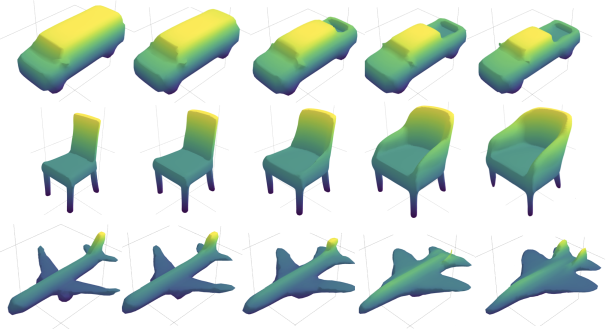}
\caption{Linear shape interpolation between objects of the same class of the ShapeNet-Core V1 test set. (row 1) car-car, (row 2) chair-chair, (row 3) table-table, (row 4) plane-plane}
\label{fig:interp}
\end{figure}

\begin{figure}[t]
\centering
\includegraphics[width=.95\linewidth]{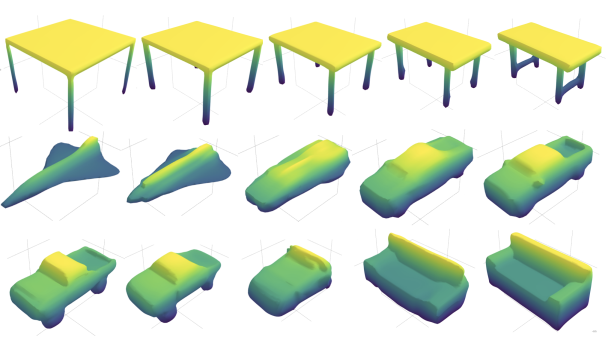}
\caption{Linear shape interpolation between objects from different classes of the ShapeNet-Core V1 test set. (row 1) table-bench, (row 2) plane-car, (row 3) car-couch}
\label{fig:interp2}
\end{figure}

As Fig.~\ref{fig:tsne} shows, the embedding $e$ obtained by the network (of size $16N$, see Sec.~\ref{sec:arch}) has learned to separate between the classes in an unsupervised way. The learned embedding also presents what can be considered a quasi-linear behavior in the semantic space. This is evident in Fig.~\ref{fig:interp}, in which the embeddings $e_1$ and $e_2$ obtained from  single image $I_1,I_2$ of two random shapes from the same class of the test set are linearly interpolated ($\lambda*e_1 + (1-\lambda)e_2$) using the interpolation weights $\lambda=0,0.25, 0.5, 0.75, \text{and}~ 1$. This effect is not limited to same class objects, and as can be seen in Fig~\ref{fig:interp2} objects from different classes also blend successfully. As far as we know, we are the only method out of the related work that presents cross-class interpolations.

\begin{figure}
\centering
\includegraphics[width=.8910\linewidth]{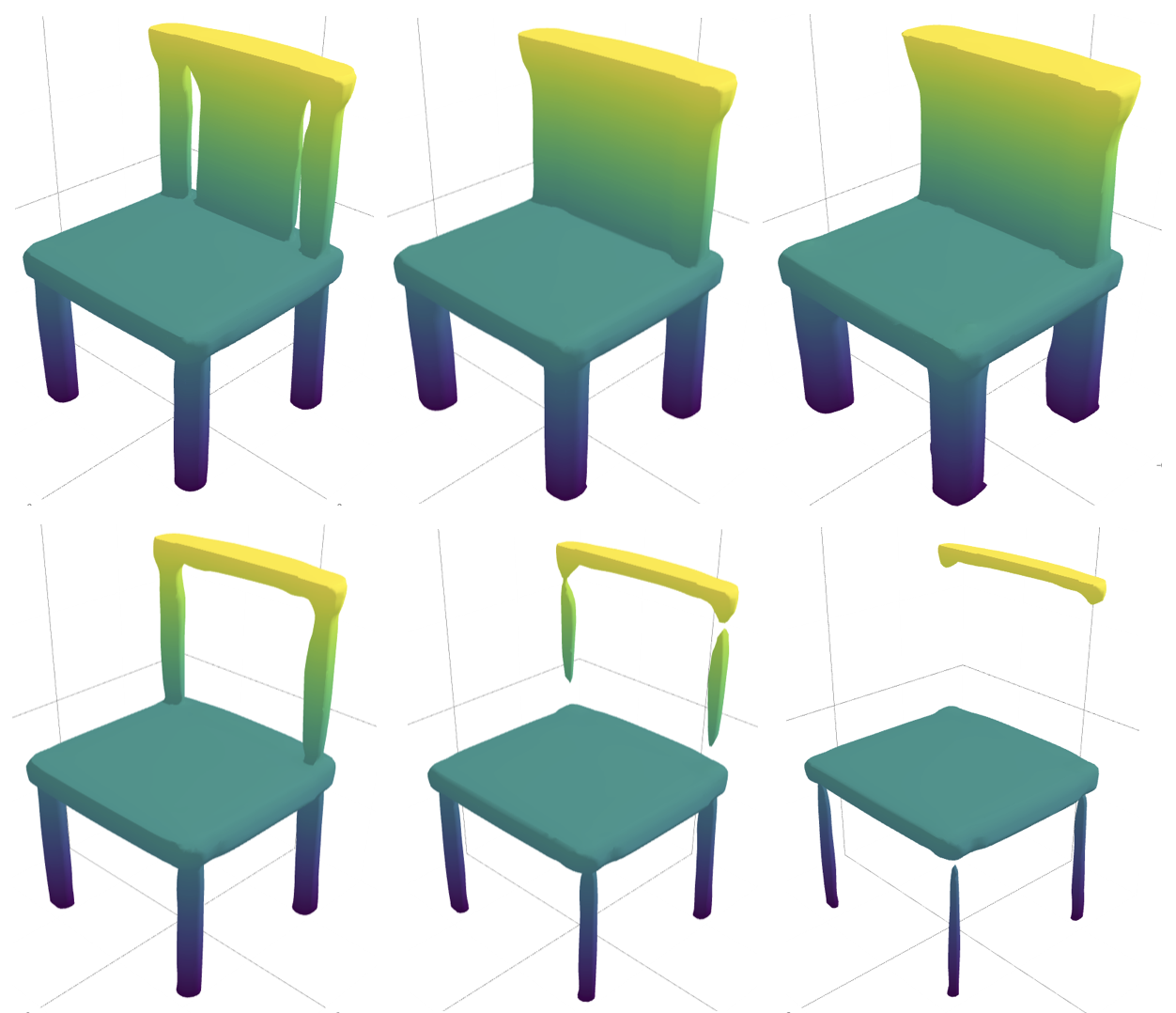}
\caption{Shape surface extracted with different thresholds on $s_I^p$, corresponding to different level-sets of the implicit field. From bottom right going clockwise: $0.9, 0.7, 0.6, 0.5, 0.3, 0.1$.}
\label{fig:threshold}
\end{figure}

The resulting scalar field $S_I^p$ encodes the object in a stable manner. When varying the threshold between 0.1 and 0.9, we obtain shapes that resemble the shape at the default 0.5 threshold, as can be seen in Fig.~\ref{fig:threshold}.

\subsection{Quantitative results}
\paragraph{$32^3$ grid resolution}
Tab.\ref{tab:results} presents a comparison with the literature methods, conducted on the data provided by Choy et al.~\cite{choy20163d}. Both per class and average results are presented. Note that all results are provided by a single model that captures all classes, and is trained without conditioning on the class and without access to out-of-scope data in the form of pre-trained models. As can be seen, our method outperforms all literature methods in mean performance. Out of the 13 categories, our method outperforms all methods in {\color{black}12 categories and PCDI~\cite{zeng2018inferring} leads in one category (firearm)}.

\begin{figure}[t]
\centering
\includegraphics[width=.91\linewidth]{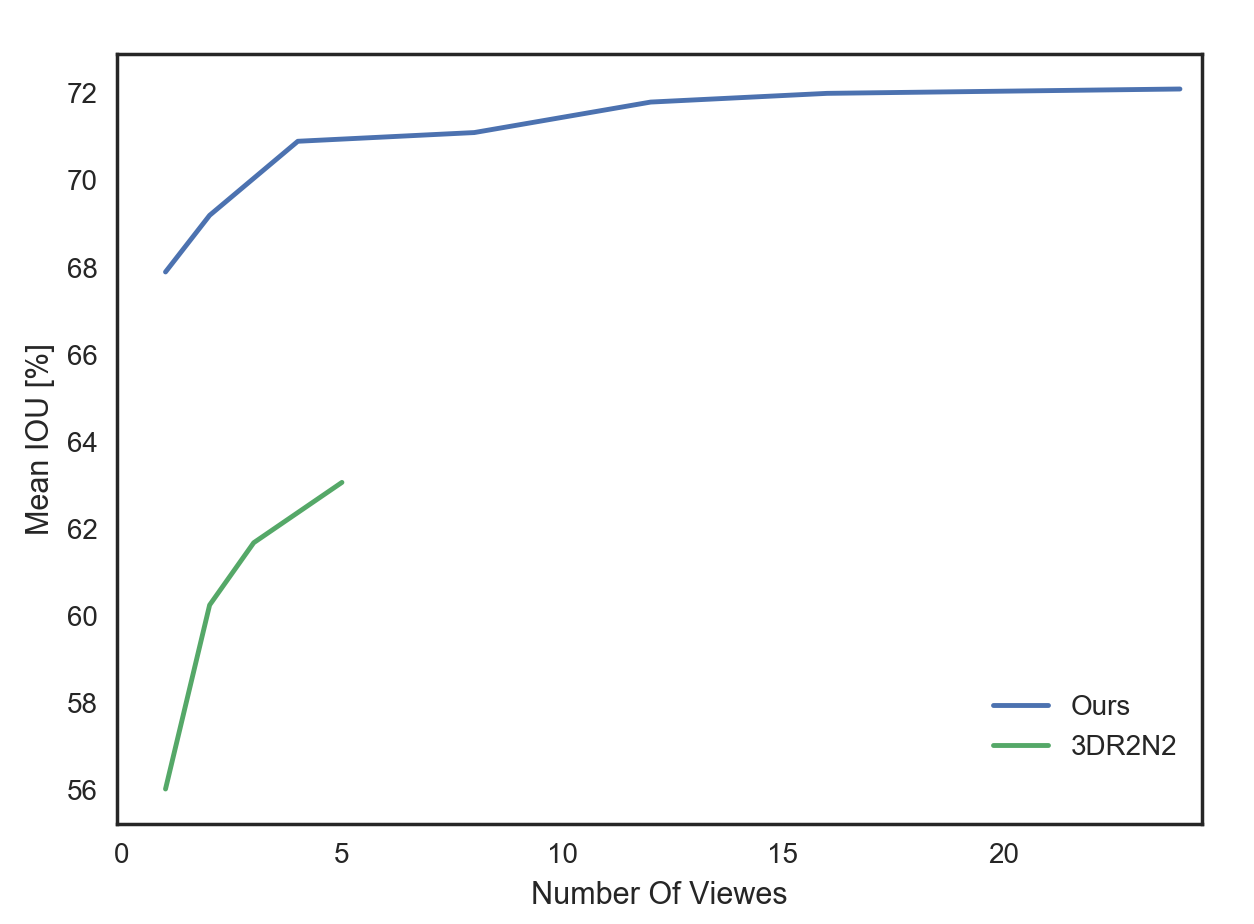}
\caption{Adding views in test-time only by averaging the embedding. The x-axis is the number of views, and the y-axis is the mean IOU. As can be seen, averaging more views improves the accuracy of the obtained shape. We compare with the reported results of 3D-R2N2~\cite{choy20163d}, which trains specifically for the multi-view scenario.}
\label{fig:multiview}
\end{figure}

In order to further evaluate the strength of our embedding, we have designed a simple multi-view test, in which during test time the embedding $e(I_i,\theta_f)$ of multiple views $I_i$ of the same shape are averaged. As can be seen in Fig.~\ref{fig:multiview}, the performance improves as the number of views increases. The ability to improve performance in a late fusion manner indicates that our embedding is well-behaved and invariant to the exact viewpoint. Our multiview results also outperform those of 3D-R2N2~\cite{choy20163d}, which is the only literature method we found to report multi-view results on the data split we employ. We stress that unlike the baseline method, we did not re-train our model to handle the multi-view task. 

\begin{figure}
\centering
\includegraphics[width=.81\linewidth]{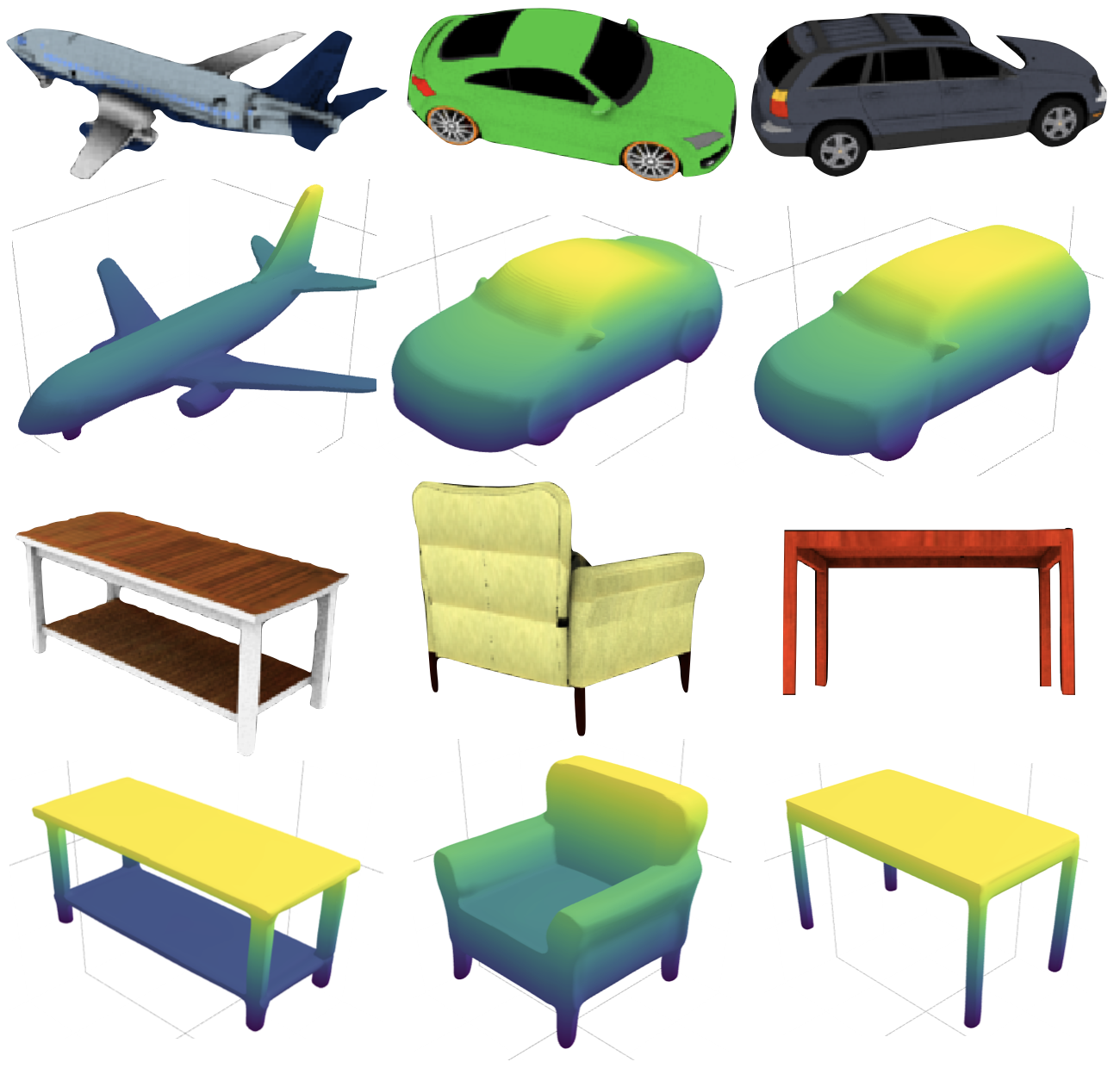}
\caption{Single Image 3D Reconstruction. (rows 1,3) input image. (rows 2,4) our results.}
\label{fig:test_data_reconstructions}
\end{figure}

\paragraph{$256^3$ grid resolution}
\textcolor{black}{Tab.\ref{tab:results2} presents a comparison with the literature methods, conducted on the data provided by H{\"a}nee et al~\cite{hane2017hierarchical}. 
In order to compare with previous work which reported results in a grid resolution of $32^3$, pooling with stride 8 was applied to the predicted voxel grid generated at test time. 
Out of the 13 categories, our method outperforms all methods in 8 categories, LSM~\cite{kar2017learning} leads in one category and VP3D~\cite{kato2018learning} leads in 4 categories. For these experiments network $g$ was parametrized by six hidden layers with 32 hidden units each $tanh$ activation was employed.  Parameters of network $f(I,\theta_f)$ were chosen as  $N=64$, $B=5$, $K=2$, and $L=4$. Although we believe IOU is a more suitable metric for task of 3D shape reconstruction, we have also evaluated our model with the Chamfer distance (CD) metric. To this end we follow the protocol and reported results provided by AtlasNet~\cite{Groueix_2018} in section 5.2 and table 4 of their publication. Results are presented in Tab.\ref{tab:results_cd} and demonstrated in Fig.~\ref{fig:test_data_reconstructions}}

\begin{table*}[t]
\begin{center}
\begin{small}
\begin{tabular}{lllllllllllllll}
\toprule
  Method & \rotatebox{90}{airplane} & \rotatebox{90}{bench} & \rotatebox{90}{cabinet} & \rotatebox{90}{car} & \rotatebox{90}{cellphone} & \rotatebox{90}{chair} & \rotatebox{90}{couch} & \rotatebox{90}{firearm} & \rotatebox{90}{lamp} & \rotatebox{90}{monitor} & \rotatebox{90}{speaker} & \rotatebox{90}{table} & \rotatebox{90}{watercraft} & \rotatebox{90}{mean} \\
\midrule
3D-R2N2~\cite{choy20163d} & 51.3 & 42.1 & 71.6 & 79.8 & 66.1 & 46.6 & 62.8 & 54.4 & 38.1 & 46.8 & 66.2 & 51.3 & 51.3 & 56.0 \\
OGN~\cite{tatarchenko2017octree} & 58.7 & 48.1 & 72.9 & 81.6 & 70.2 & 48.3 & 64.6 & 59.3 & 39.8 & 50.2 & 63.7 & 53.6 & 63.2 & 59.6 \\
PSGN~\cite{Fan_2017_CVPR} & 60.1 & 55.0 & 77.1 & 83.1 & 74.9 & 54.4 & 70.8 & 60.4 & 46.2 & 55.2 & 73.7 & 60.6 & 61.1 & 64.0 \\
VTN~\cite{richter2018matryoshka} & 67.1 & 63.7 & 76.7 & 82.1 & 74.2 & 55.0 & 69.0 & 62.6 & 43.6 & 53.4 & 68.1 & 57.3 & 59.9 & 64.1 \\
MTN~\cite{richter2018matryoshka} & 64.7 & 57.7 & 77.6 & 85.0 & 75.6 & 54.7 & 68.1 & 61.6 & 40.8 & 53.2 & 70.1 & 57.3 & 59.1 & 63.5 \\
PCDI~\cite{zeng2018inferring} & 61.2 & 60.9 & 68.3 & 83.2 & 74.4 & 57.2 & 69.9 & {\bf 69.5} & 46.4 & 61.4 & 69.8 & 61.5 & 58.5 & 64.8 \\
Ours & {\bf 71.4} & {\bf 65.9} & {\bf 79.3} & {\bf 87.1} & {\bf 79.1} & {\bf 60.7} & {\bf 74.8} & 68.0 & {\bf 48.6} & {\bf 61.7} & {\bf 73.8} & {\bf 62.8} & {\bf 65.4} & {\bf 69.1}\\
\bottomrule
\end{tabular}
\end{small}
\end{center}
\caption{Shape reconstruction from a single image on ShapeNet-core at $32^3$
grid resolution. Mean IOU (\%) per category is reported as well as the average IOU (\%) over all 13 categories. Dataset provided by Choy et al.~\cite{choy20163d}}
\label{tab:results}
\end{table*}

\begin{table*}[t]
\begin{center}
\begin{small}
\begin{tabular}{lllllllllllllll}
\toprule
  Method & \rotatebox{90}{airplane} & \rotatebox{90}{bench} & \rotatebox{90}{cabinet} & \rotatebox{90}{car} & \rotatebox{90}{cellphone} & \rotatebox{90}{chair} & \rotatebox{90}{couch} & \rotatebox{90}{firearm} & \rotatebox{90}{lamp} & \rotatebox{90}{monitor} & \rotatebox{90}{speaker} & \rotatebox{90}{table} & \rotatebox{90}{watercraft} & \rotatebox{90}{mean} \\
\midrule
3D-R2N2~\cite{choy20163d} & 56.7 & 43.2 & 61.8 & 77.6 & 65.8 & 50.9 & 58.9 & 56.5 & 40.0 & 44.0 & 56.7 & 51.6 & 53.1 & 55.1 \\
LSM~\cite{kar2017learning} & 61.1 & 50.8 & 65.9 & 79.3 & 67.7 & 57.8 & 67.0 & {\bf 69.7} & 48.1 & 53.9 & 63.9 & 55.6 & 58.3 & 61.5 \\
VP3D~\cite{kato2018learning} & 69.1 & 59.8 & 72.4 & 80.2 & {\bf 77.5} & 60.1 & 65.6 & 66.4 & 50.5 & {\bf 59.7} & {\bf 68.0} & 60.7 & {\bf 61.3} & 65.5 \\
Ours & {\bf 71.3} & {\bf 63.4} & {\bf 75.6} & {\bf 81.5} & 75.1 & {\bf 61.4} & {\bf 72.3} & 65.7 & {\bf 52.0} & 56.2 &  64.7 & {\bf 61.6} & 60.2 & {\bf 66.2}\\
\bottomrule
\end{tabular}
\end{small}
\end{center}
\caption{Same as Tab.~\ref{tab:results} for the dataset provided by H{\"a}nee et al~\cite{hane2017hierarchical}.}
\label{tab:results2}
\end{table*}

\begin{table}[t]
\centering
\begin{tabular}{lccc}
\toprule
   & HSP~\cite{hane2017hierarchical} & AtlasNet~\cite{Groueix_2018} & Ours\\
  Average CD $\times 10^3$  & 11.6& 9.52&{\bf 4.35}\\
\bottomrule
\end{tabular}
\caption{Shape reconstruction from a single image on ShapeNet-core at $256^3$
grid resolution. Average CD (\%) is reported over all 13 categories. Dataset provided by H{\"a}nee et al~\cite{hane2017hierarchical}. The Chamfer Distance (CD) reported is computed on 10000 uniformly sampled points, multiplied by $10^ 3$ and averaged over all classes.}
\label{tab:results_cd}
\end{table}

\subsection{Parameter sensitivity}

Since only one loss term is used, there are not many parameters to select, except for the architecture of the network itself. The method seems to be insensitive to the selection of architecture. In Tab.~\ref{tab:parametersensitivityg}, we evaluated the sensitivity of the method to the architecture of the network $g$ used to represent each shape. These experiments were run for 12 epochs and not until convergence. As can be seen, the performance is relatively constant across the three activation functions tested (ELU, ReLU, and $tanh$) and for a wide range of the number of layers and number of hidden units per layer.

Sensitivity was also evaluated with respect to the parameters of network $f$. To that end, we tested four different ResNet architectures. We parameterize them by the number of blocks $(B\in\{4,5\})$, number of base kernels $(N\in\{64,128\})$ and number of fully connected layers $(K\in\{0,2\})$. Overall, it seems that there is little sensitivity to the parameters and a slight preference to the larger number of blocks $B=5$.

\begin{table}
\begin{center}
\begin{small}
\begin{tabular}{@{}lc@{~~}c@{~~}cc@{~~}c@{~~}cc@{~~}c@{~~}c@{}}
\toprule
& \multicolumn{3}{c}{ELU} & \multicolumn{3}{c}{ReLU} & \multicolumn{3}{c}{$tanh$}\\
\cmidrule(lr){2-4} \cmidrule(lr){5-7} \cmidrule(lr){8-10}
 & 16 & 32 & 64& 16 & 32 & 64& 16 & 32 & 64\\
 \midrule
3 & 65.2 & 65.2 & 66.1  & 65.4 & 65.6 & 66.1 & 65.1 & 65.4 & 65.7 \\
4 & 65.4 & 65.6 & 65.8  & 65.1 & 65.5 & 66.0 & 65.8 & 64.9 & 66.1 \\ 
5 & 64.8 & 65.5 & 66.1  & 65.2 & 65.5 & 65.9 & 65.6 & 65.7 & 65.3 \\
6 & 65.7 & 65.5 & 66.0  & 64.5 & 65.1 & 65.8 & 64.8 & 65.4 & 65.6 \\ 
\bottomrule
\end{tabular}
\end{small}
\end{center}
\caption{Sensitivity to the hyperparameters of $g$. Reported is the IOU (\%)  after 12 epochs for a network trained with the ELU, ReLU, or $tanh$ activation. 
Each row (column) has a different number of layers (hidden units per layer).}
\label{tab:parametersensitivityg}
\end{table}
\begin{table}
\begin{center}
\begin{small}
\begin{tabular}{lcccc}
\toprule
\multicolumn{3}{c}{Hyperparameters}  & IOU\\
\cmidrule{1-3}
N & B & K &  \\
 \midrule
64 & 4 & 2 & 67.0 \\
128 & 4 & 0 & 67.0 \\
64 & 5 & 2 & 67.3 \\
128 & 5 & 0 & 67.3\\
\bottomrule
\end{tabular}
\end{small}
\end{center}
\caption{Sensitivity to parameters of network $f$. Reported is the IOU (\%) after convergence for a network trained with different settings of hyperparameters. Experiments were run for 15 epochs.}
\label{tab:parametersensitivityf}
\end{table}

\subsection{Sampling Technique}

We evaluate our sampling method by comparing the accuracy over epochs, obtained  by a network that was trained with boundary sampling versus a network that was trained with random uniform sampling in the $[-1,1]$ volumetric cube. For a fair comparison, both networks share the same relatively lightweight architecture ($N=64$,$B=5$,$K=2$) and were trained with the same set of hyper-parameters for 20 epochs, without lowering the learning rate. The network trained with boundary sampling reached a mean IOU score of $65.8\%$ vs. $63.5\%$ for the network trained with random uniform sampling.

\begin{figure}[t]
\centering
\includegraphics[width=0.91\linewidth]{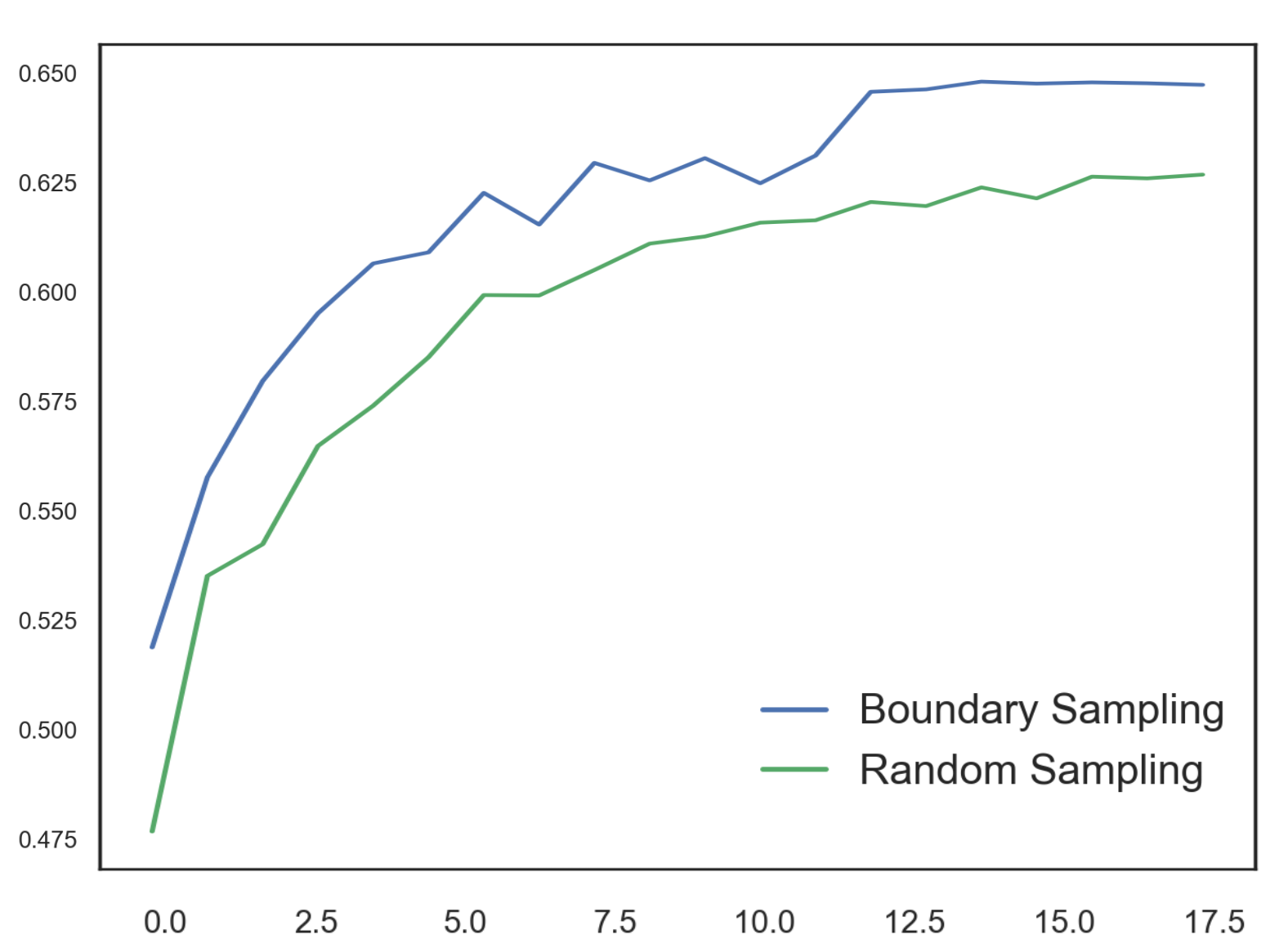}
\caption{Learning curves for ShapeNet showing mean IOU vs. training epoch on the test set. Training with boundary sampling (blue) is compared to training with random uniform sampling (green). The learning rate was not reduced in these runs, in order not to bias the results toward the timing of a specific scenario.}
\label{fig:sampling_study}
\end{figure}

\subsection{Reconstruction of real-world images}
\begin{figure}
\centering
\includegraphics[width=.971\linewidth]{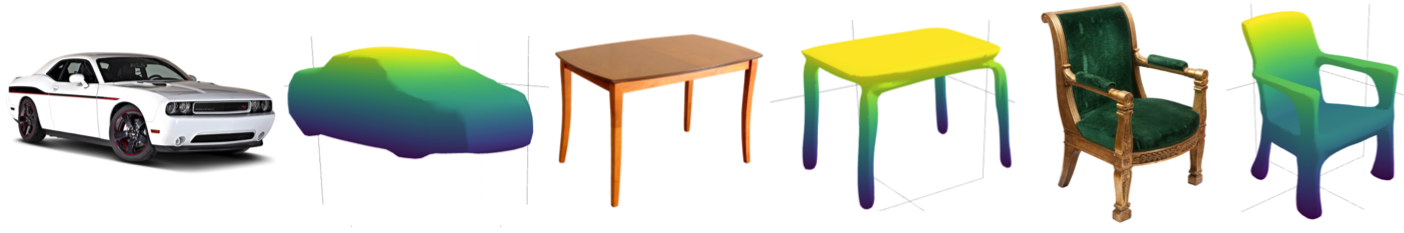}
\caption{Reconstruction from real-world images. (left) input image. (right) reconstruction result.}
\label{fig:realworld}
\end{figure}

We follow previous methods and test our model on real-world images from the internet, using the same model trained on the ShapeNet dataset. As demonstrated in Fig.~\ref{fig:realworld}, our model generalizes well across different categories. However, we notice that successful reconstruction is dependent on the point of view. Since the existing datasets are very biased in that respect, a next step would be to render a more uniformly distributed dataset with respect to camera parameters.  

\subsection{Visualization of Jacobian norm}
We wish to evaluate the gradients of $g$ with respect to $p(x,y,z)$ which correspond to the Jacobian of $g$.
\begin{equation}
J(\theta_f,I,p) =  \frac{\partial(g(p,f(I,\theta_f)))}{\partial p} 
\end{equation}
In order to evaluate the traits of the shape boundary, we calculate the Jacobian norm at the zero levelset: $\left|J(\theta_f,I,p)\vert_{s(p)=0}\right|$.

\begin{figure}
\centering
\includegraphics[width=.81\linewidth]{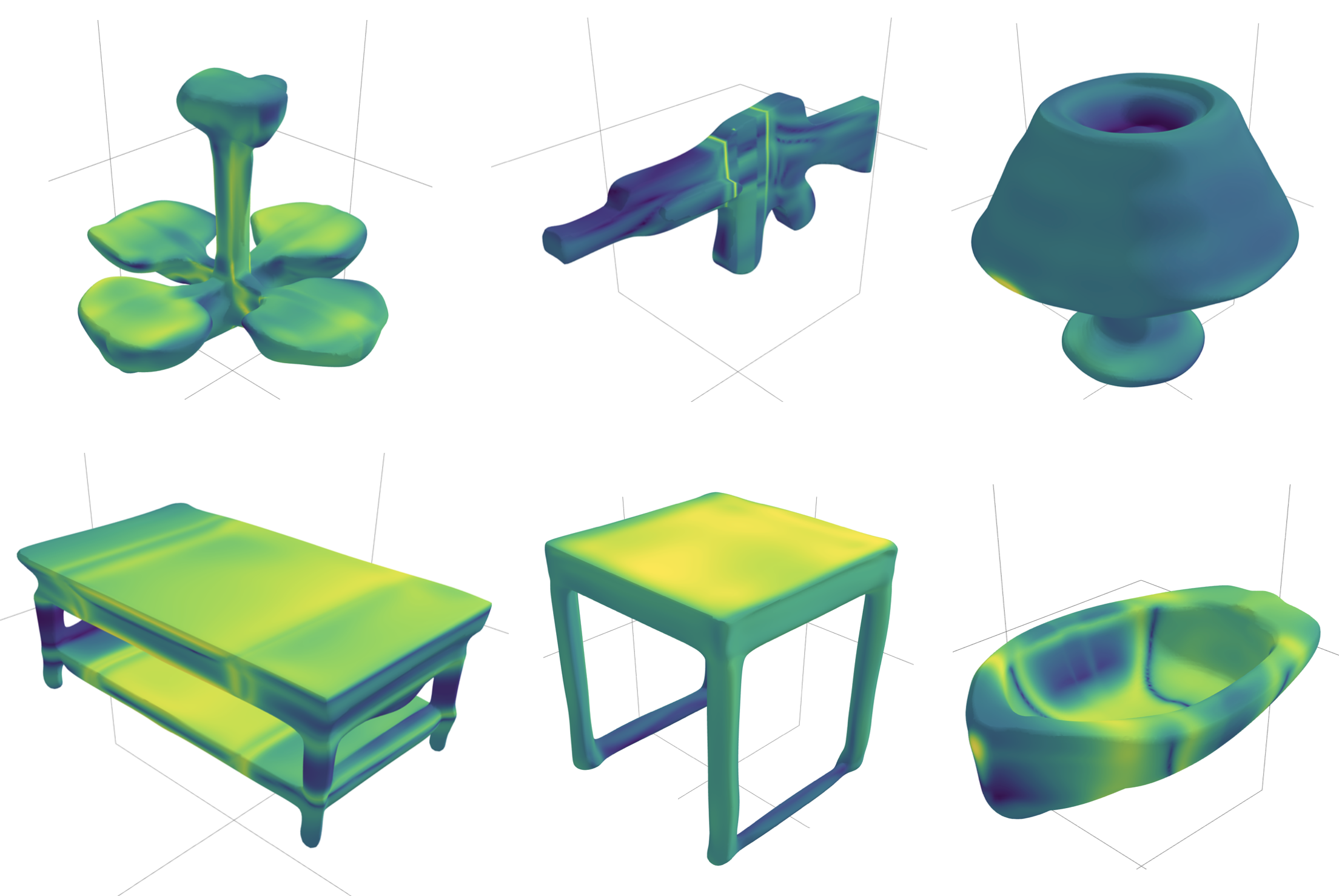}
\caption{Jacobian norm values evaluated on the shape surface. }
\label{fig:gradient_norm}
\end{figure}

Ths obtained norm can be viewed as a local sensitivity score of the shape, or as some sort of confidence. It is displayed in Fig.~\ref{fig:gradient_norm} on a scale in which the low norms are yellow and the high norms are bluish. Flat surfaces present smaller gradient norms than the highly curved surfaces. Note that the direction of the gradient is always normal to the surface, which is a property of isosurfaces.

\section{Extensions}
\label{sec:discussion}

The simplicity of our method in comparison to the alternative representations leads to straightforward extensions. For example, we can model dynamic shapes simply by employing functionals $g$ with inputs of the form $p=(x,y,z,t)$, where $t$ represents the time dimension, and recover the weights of $g$ using a learned function $f$, which takes a sequence of images as input. A loss term $S$ could be added to encourage $g$ to be smooth in time near the model's boundary:
\begin{equation}
S(\theta_f,I) = -\int_{V\times T} \left| \frac{\partial(g(p,f(I,\theta_f)))}{\partial t} \right|dp 
\end{equation}
where $I$ is now a sequence of images, $y$ the ground truth sequence of 3D shapes, $V$ is the 3D volume and T is the time dimension. Note that this extension requires very little change to the method's code. In comparison, if one were to model sequences in time using meshes or voxels, the added complexity of the representation would be significant, making high resolution models less tractable, and the smoothness over time would require significant code.

\begin{figure}
\centering
\includegraphics[width=.91\linewidth]{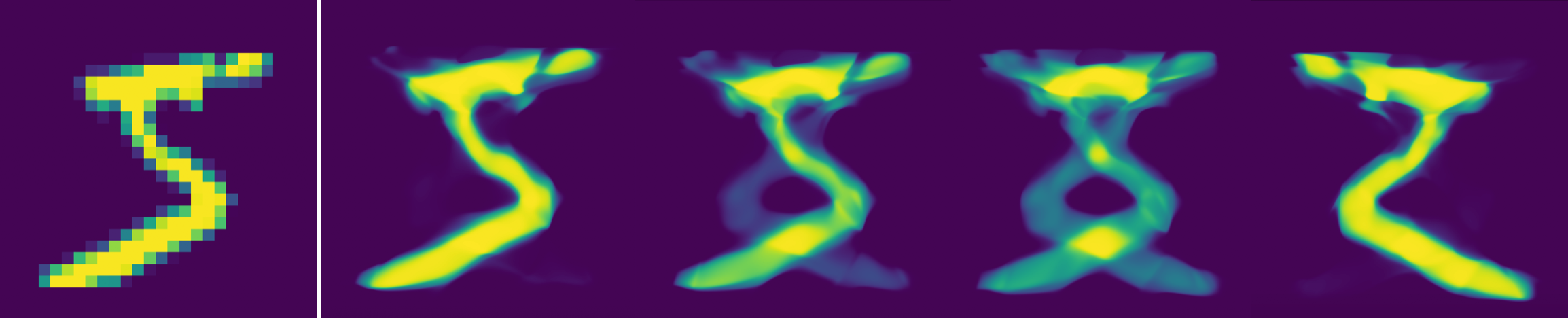}
\caption{A sequence of temporal reconstruction results obtained from the single digit on the left.}%
\label{fig:temporal}
\end{figure}

In a POC experiment, we autoencode the mnist dataset where $g$ maps a 3D point $p$ with coordinates $(x,y,\alpha)$ to a value in $[0,1]$. $\alpha$ is a dynamic parameter which smoothly interpolates between a digit and its mirrored version. As can be seen in Fig.~\ref{fig:temporal}, from a single view $(28^2)$, the method learns to generate the entire sequence (rendered for different values of $\alpha$ and at a higher resolution of $1024^2$).

The method can be also applied directly beyond points to other geometric primitives. For example, the functional can indicate whether a set of three points is a triangular mesh that belongs to a shape's boundary or not.

Our representation also opens up interesting options in the realm of differentiable rendering. Implicit fields have long been used for graphical applications \cite{bloomenthal1988polygonization}. Several rendering techniques, such as ray tracing \cite{hanrahan1983ray} and sphere tracing \cite{article} were designed to deal with the task of projecting these fields into 2D in order to generate images. Since our inferred implicit field is differentiable anywhere, applying these techniques results in the ability to back-propagate errors generated by image-image comparison. This could lead, for example, to efficient multi-image training.

A POC implementation of a differentiable renderer was conducted in the context of learning 3D from silhouettes using a simple L2 loss with the ground truth silhouettes, which were captured from three canonical viewpoints around the object. Silhouettes were rendered by max pooling $sigmoid(g(p))$ for points $p$ along the 3D rays associated with every image pixel. We obtain an IOU of 64.4, whereas the literature for learning from three silhouettes~\cite{kato2018learning} gets 60.0. However, the three views used were different.

\section{Conclusion}

Learning the novel functional representation of shapes introduced in this work requires only a single loss term. The smooth manifold obtained has a high capacity. The method is elegant, simple to implement, and easily extendable. The embedding learned by the method displays an intuitive semantic behavior and averaging  in this latent space, multiple representations obtained from different views leads to more accurate shapes. Our experiments indicate that the new representation leads to more accurate results than the literature methods for the task of 3D reconstruction from a single view.

\section*{Acknowledgement}
This project has received funding from the European Research Council (ERC) under the European Unions Horizon 2020 research and innovation programme (grant ERC CoG
725974).

{\small

}

\end{document}